\documentclass[letterpaper,times]{IONconf}
\usepackage{amsmath,amssymb,amsfonts}
\usepackage[hidelinks]{hyperref}

\usepackage{url}
\usepackage{amsmath}
\usepackage{cite}
\usepackage{subcaption}
\usepackage[pdftex]{graphicx}
\usepackage{authblk}
\usepackage{array}
\usepackage[flushleft]{threeparttable}
\usepackage[symbol]{footmisc}

\usepackage{titlesec}
\titleformat{\section}
  {\normalfont\bfseries\uppercase}{\thesection.}{0.5em}{}
\renewcommand\thesection{\Roman{section}}

\title{ZUPT aided GNSS Factor Graph with Inertial Navigation Integration for Wheeled Robots}

\author[1]{Cagri Kilic\thanks{}}
\author[1]{Shounak Das\footnote[1]{These authors contributed equally to this study.}}
\author[1]{Eduardo Gutierrez}
\author[2]{Ryan Watson}
\author[1]{Jason Gross}
\affil[1]{Department of Mechanical and Aerospace Engineering \\
West Virginia University \\
Morgantown, WV 26505, USA }
\affil[2]{The Johns Hopkins University Applied Physics Laboratory, Laurel, USA}

\usepackage{acronym}

    \acrodef{WVU}{West Virginia University}
    \acrodef{MAE}{Mechanical and Aerospace Engineering}

    \acrodef{IMU}{inertial measurement unit}
    \acrodef{INS}{inertial navigation system}
    \acrodef{GPS}{global positioning system}
    \acrodef{GNSS}{global navigation satellite system}
    \acrodef{LiDAR}{Light detection and ranging}
    \acrodef{ZUPT}{zero velocity update}
    \acrodef{WO}{wheel odometry}
    
    \acrodef{PLL}{phase lock loop}
    \acrodef{DLL}{delay lock loop}
    \acrodef{IQ}{in-phase and quadrature}
    \acrodef{SDR}{software defined radio}
    \acrodef{RINEX}{receiver independent exchange format}
    \acrodef{RTK}{real time kinematic}
    \acrodef{PPP}{precise point positioning}
    
    \acrodef{SQM}{signal quality metric}
    \acrodefplural{SQM}[SQMs]{signal quality metrics}
    \acrodef{SS}{signal strength}
    \acrodef{EL}{elevation angle}
    \acrodef{AZ}{azimuth angle}

    \acrodef{KF}{Kalman filter}
    \acrodef{EKF}{extended Kalman filter}
    \acrodef{iSAM2}{incremental smoothing and mapping}
    \acrodef{GMM}{Gaussian mixture model}
    \acrodef{LS}{least squares}
    \acrodef{NLLS}{nonlinear least squares}
    \acrodef{SLAM}{simultaneous localization and mapping}
    \acrodef{MAP}{maximum a posteriori}
    \acrodef{LARS}{least angel regression}
    
    \acrodef{m-estimator}{maximum likelihood type estimator}
    \acrodefplural{m-estimator}[m-estimators]{maximum likelihood type estimators}
    
    \acrodef{IRLS}{iteratively re-weighted least squares}
    \acrodef{DCS}{dynamic covariance scaling}
    \acrodef{MM}{max-mixtures}
    \acrodef{BCE}{batch covariance estimation}
    \acrodef{ICE}{incremental covariance estimation}
    \acrodef{BCE-AD}{batch covariance estimation over an augmented data-space}
    \acrodef{RANSAC}{random sample consensus}
    \acrodef{RRR}{realizing, reversing, recovering}
    \acrodef{RAIM}{receiver autonomous integrity monitoring}
    \acrodef{SCGP}{single-cluster spectral graph partitioning}

    \acrodef{MAP}{maximum a posteriori}
    \acrodef{MLE}{maximum likelihood estimate}

    \acrodef{RSOS}{residual-sum-of-squares}
    \acrodef{HRSOS}{horizontal residual-sum-of-squares}
    \acrodef{TRSOS}{total residual-sum-of-squares}
    \acrodef{RMSE}{root mean square error}
    
\acrodef{NN}{nearest neighbor}
\acrodef{VDP}{variational Dirichlet process}
\acrodef{VI}{variational inference}

\usepackage[pscoord]{eso-pic}
\newcommand{\placetextbox}[3]{
\setbox0=\hbox{#3}
\AddToShipoutPictureFG{ \put(\LenToUnit{#1\paperwidth},\LenToUnit{#2\paperheight}){\vtop{{\null}\makebox[0pt][c]{#3}}}}
}
\placetextbox{.5}{0.05}{Preprint Version. Published in: ION GNSS+ 2021, St. Louis, Missouri, September 2021, pp. 3285-3293.} 
\placetextbox{.5}{0.035}{ 
\url{https://doi.org/10.33012/2021.18064}}
\begin{document}

\maketitle

\section*{Abstract}

In this work, we demonstrate the importance of zero velocity information for \ac{GNSS} based navigation. The effectiveness of using the zero velocity information with \ac{ZUPT} for inertial navigation applications have been shown in the literature. Here we leverage this information and add it as a position constraint in a \ac{GNSS} factor graph. We also compare its performance to a \ac{GNSS}/\ac{INS} coupled factor graph. We tested our \ac{ZUPT} aided factor graph method on three datasets and compared it with the GNSS-only factor graph.

\section{Introduction}

Localization is an integral component of any mobile robot system, which plays an important role in many core robotic capabilities like motion planning, obstacle avoidance, and mapping. One of the most common methods for calculating positioning information of the robots is using \ac{GNSS}~\cite{enge1994global}. However, the availability of this system relies on the observation of multiple satellites~\cite{grovebook} and is frequently unable to obtain a precise and robust state estimate in urban and forested areas due to environmental constraints (e.g., poor satellite geometry and multipath effects)~\cite{merry2019smartphone}. An \ac{IMU} can be used in a standalone manner in the \ac{INS} to estimate states but suffers from drifting due to the accumulation of errors through the integration of acceleration and angular velocity measurements \cite{kok2017using,chen2018ionet,brossard2020ai,narasimhappa2019mems}.
A commonly adopted localization strategy is coupling \ac{GNSS} and \ac{INS}, which combines range measurements from satellites and IMU measurements to calculate states~\cite{zhao2016analysis,li2006low,miller2012sensitivity,hu2015derivative}. This coupling strategy alleviates some of the vulnerabilities that standalone \ac{GNSS} technology faces in urban environments and forested areas~\cite{merry2019smartphone}. 

Though Kalman filters have been the preferred choice for \ac{GNSS} and \ac{GNSS}/\ac{INS} state estimation for a long time, factor graphs~\cite{dellaert2017factor} have emerged as an alternative framework for solving the localization problem mostly because of the advent of open-source graph optimization libraries like GTSAM~\cite{dellaert2012factor}, g2o~\cite{grisetti2011g2o}. Factor graph optimization has some advantages over the Kalman filters. For example, it uses multiple iterations to solve all states in a batch form instead of just one iteration in a Kalman filter, which is not affected by future measurements (unless smoothing is done). Factor graphs have also been seen to explore better the time correlation between past and current epochs, which helps in outlier removal~\cite{wen2021factor}. The factor graph framework also makes it easy to add existing and new robust estimation methods~ \cite{watson2018evaluation,watson2020robust,pfeifer2018robust} which helps reduce localization failure due to large noise.

In order to provide a cost-effective system, one can utilize the pseudo-measurements in certain conditions from the sensors already on-board. For instance, \ac{ZUPT} is commonly used as an aiding process for pedestrian navigation~\cite{kwakkel2008gnss,zhang2017adaptive}. \ac{ZUPT} can bound the velocity error and help to calibrate \ac{IMU} sensor noises~\cite{skog2010zero}. This process helps to reduce the \ac{INS} positioning error growth from cubic to linear since the error state model justifies the correlation between the position and velocity errors of the error covariance matrix. Using \ac{ZUPT} in state estimation does not need any dedicated sensor to provide zero velocity information, and this information can be obtained by the sensors already on-board (e.g., \ac{IMU}, wheel encoders). \ac{ZUPT} requires stationary conditions, and it can be used as an opportunistic navigational update if a wheeled robot stops for other reasons (e.g., obstacle avoidance, re-planning, waiting for pedestrians, stopping at traffic lights). Also, \ac{ZUPT} can be used to improve \ac{WO}/\ac{INS} proprioceptive localization with periodic stops in \ac{GNSS}-denied~(or degraded), poor lighting/feature areas~\cite{kilic2019improved} and with autonomous stops by deciding when to stop~\cite{kilic2020}. The small wheeled robots have more freedom of stopping rather than autonomous cars, which makes utilizing \ac{ZUPT} a well-suited application for them.

This paper offers the following contributions: 1) detailing a \ac{GNSS}/\ac{WO}/\ac{INS} fusion strategy by exploiting the zero velocity information for both \ac{GNSS} and \ac{INS} to be used for wheeled robots, 2) validation of the provided method on actual hardware in field tests with detailed specifications of the implementation and hardware, so the interested reader can easily replicate our work, and 3) making our algorithm publicly available\footnote{\url{https://github.com/wvu-navLab/gnss-ins-zupt}} using open-access data~\cite{vz7z-jc84-20}. 

The rest of this paper is organized as follows. Section~\ref{method} details and explains the components of the algorithm. Section~\ref{experimental_results} provides the experimental results. Finally, Section~\ref{conclusion} provides a conclusion and insights for future works to improve the approach.

\section{Methodology}
\label{method}

\subsection{GNSS Factor Graph}
\label{FG}

\begin{figure}[h]
\centering
\includegraphics[width=0.85\linewidth]{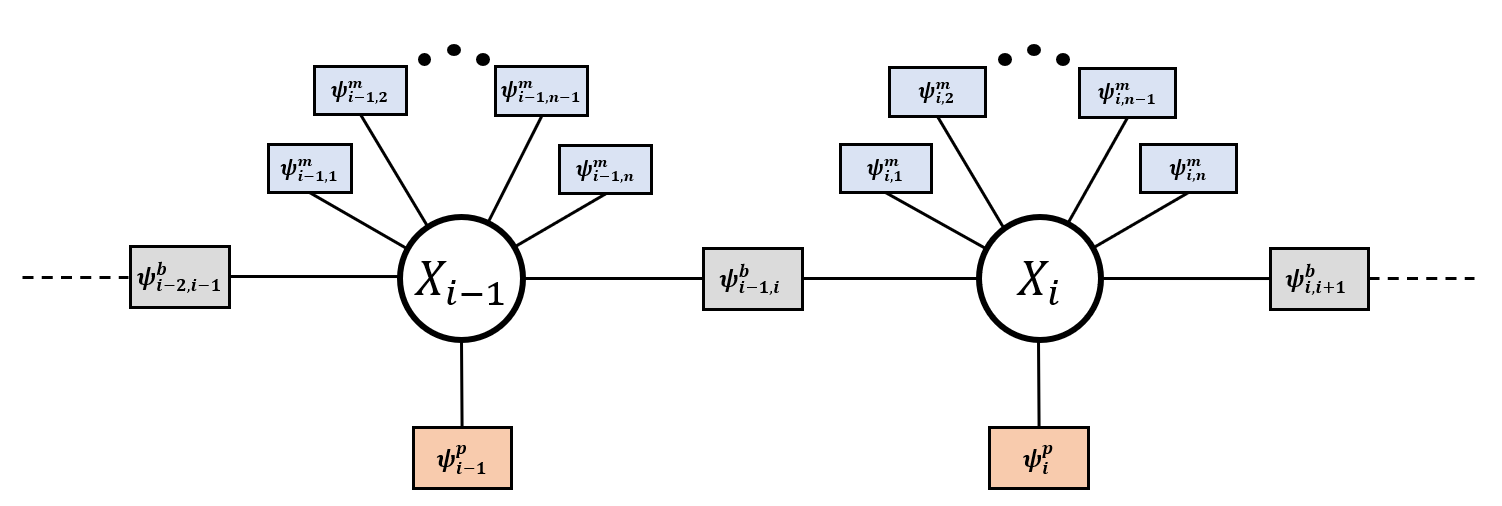}
\caption{GNSS factor graph example. }
\label{fig:factorgraph}
\end{figure}

The state estimation framework used for processing \ac{GNSS} data is a factor graph~\cite{dellaert2017factor}. Factor graphs solve the \ac{MAP} estimation problem by maximizing the product of factors that are probabilistic constraints between states at different time steps and between states and measurements from the current time steps. A depiction of an example factor graph is given in~Fig \ref{fig:factorgraph}.

A detailed description of creating factors with \ac{GNSS} observations is presented in~\cite{watson2018evaluation}. The states estimated in the GNSS factor graph are the position, tropospheric delay, clock bias, and phase bias. In Fig.~\ref{fig:factorgraph}, $\psi$ represents any constraint that might exist between the states and the measurements. For example, $\psi^p$ represents any prior belief on each state, $\psi^b$ is the motion constraint between two consecutive states along the trajectory, and $\psi^m$ is the measurement constraint between a state, where the measurements that were perceived from that state. Under the Gaussian assumption, the maximum product of the factors problem converts to a non-linear least squares problem. Following GTSAM nomenclature, we refer to $\psi^b$ as the between-factor. The cost form of the factors is shown in the equation~\ref{eq:factorcost}. Each component of the sum is a Mahalanobis cost, the left one being the prior factor cost, the middle one being the between-factor cost($\Delta$ is the measured displacement), and the right one is the GNSS factor cost.

\begin{equation}
\hat{X}=\underset{x}{\operatorname{argmax}}\left\{\prod_{i=1}^{I} \psi_i^p \prod_{j=1}^{J} \psi_{j-1,j}^b \prod_{k=1}^{K} \psi_k^m\right\}
\end{equation}

\begin{equation}
\label{eq:factorcost}
\hat{X}=\underset{x}{\operatorname{argmin}}\left[\sum_{i=1}^{I}\left\|x_{o}-x_{i}\right\|_{\Sigma}^{2}+\sum_{j=1}^{J}\left\|(x_{j}-x_{j-1})-\Delta\right\|_{\Lambda}^{2}+\sum_{k=1}^{K}\left\|z_{k}-h_{k}\left(x_{k}\right)\right\|_{\Xi}^{2}\right]
\end{equation}

As new \ac{GNSS} measurements are received, new factors are added using pseudo-range and phase measurements to the existing factor graph and solved incrementally for every epoch using the Incremental Smoothing and Mapping~(iSAM2) formulation~\cite{kaess2012isam2}. It converts the factor graph to the Bayes tree's data structure, whose vertices represent cliques for efficient inference. Thus, adding a new constraint to the factor graph only results in the re-linearization of the states within the specific clique.

\subsection{INS/WO Sensor Fusion (CoreNav)}
\label{corenav}

\begin{figure}[h]
\centering
\includegraphics[width=0.78\linewidth]{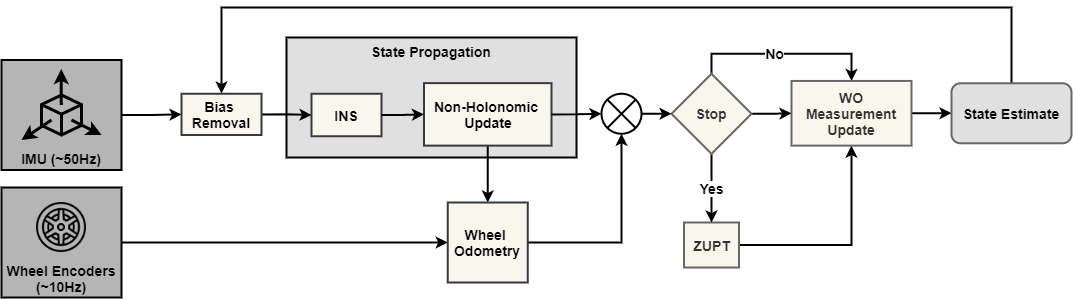}
\caption{The architecture of the \ac{INS}/\ac{WO} sensor fusion (CoreNav) framework. The state error is propagated with \ac{IMU} and corrected with \ac{WO} measurements in an error-state \ac{EKF} framework. \ac{ZUPT} and non-holonomic update are used to provide valuable information to calibrate the \ac{IMU} biases. This is done when a \ac{ZUPT} is triggered with stationary conditions whereas the non-holonomic update is performed for each state. }
\label{fig:corenav}
\end{figure}

The method in~\cite{kilic2019improved} is adopted to improve the proprioceptive localization framework, an error-state \ac{EKF} that uses IMU for state propagation and \ac{WO} for state correction. It also uses \ac{ZUPT} and non-holonomic constraints as pseudo-measurements to reduce the solution drift. In this \ac{INS}/\ac{WO} sensor fusion framework, it is assumed that a small four-wheeled robot uses a set of sensors, including a \ac{GNSS} receiver and antenna, an \ac{IMU}, and wheel encoders. The architecture of the \ac{INS}/\ac{WO} sensor fusion is provided in Fig.~\ref{fig:corenav}. This sensor fusion method is referred to as CoreNav in this paper.

Following the formulation based on~\cite{grovebook,kilic2019improved}, the error state vector can be constructed in a local navigation frame, 
\begin{equation}
\label{errorstate}
\mathbf{x}_{err}^{n}={\biggl(
\delta\mathbf{ \Psi}_{nb}^{n} \ \ 
\mathbf{\delta v}_{eb}^{n} \ \ 
\delta\mathbf{p}_{b} \ \ 
\mathbf{b}_a \ \ 
\mathbf{b}_g
\biggr )}^{\mathbf{T}}
\end{equation}
where, $\delta\mathbf{ \Psi}_{nb}^{n}$ is the attitude error, $\mathbf{\delta v}_{eb}^{n}$ is the velocity error, $\delta\mathbf{p}_{b}$ is the position error, $\mathbf{b}_a$ is the \ac{IMU} acceleration bias, and $\mathbf{b}_g$ is the \ac{IMU} gyroscope bias. For the sake of completeness, the measurement innovations for \ac{ZUPT}, non-holonomic update, and \ac{WO} are provided in this section. The detailed derivations and implementation details can be found in~\cite{grovebook,kilic2019improved}.

\subsubsection{Zero Velocity Update}
\ac{ZUPT} can be used in three ways for wheeled robots: 1) passive \ac{ZUPT}, as an opportunistic navigational update when rover needs to stop for various reasons, 2) active \ac{ZUPT}, with periodic stopping~\cite{kilic2019improved}, and 3) active \ac{ZUPT}, by deciding when to stop autonomously~\cite{kilic2020}.
Assuming a stationary rover has zero velocity and zero angular rate (i.e., constant heading), the measurement innovation for a \ac{ZUPT} can be given as  
 \begin{equation}
 \mathbf{\delta z}_{Z,k}^{n }=[-\mathbf{\hat{v}}_{eb}^{n} -\mathbf{\hat{\omega}}_{ib}^{b}]_{k}^{ T} 
 \end{equation}
where $\mathbf{\delta z}_{Z,k}^{n -}$ is measurement innovation, $\mathbf{\hat{v}}_{eb}^{n} $ is estimated velocity vector, and $\mathbf{\hat{\omega}}_{ib}^{b}$ estimated gyro bias.

\subsubsection{Non-Holonomic Update}
If a wheeled-robot cannot move sideways with its wheels, then the rover's velocity is zero along the rotation axis of its wheels, assuming it does not experience side-slip. Additionally, wheeled-robot cannot move perpendicular to the traversal surface. For this reason, these motion constraints can be used as non-holonomic pseudo-measurement update in the filter. Non-holonomic measurement innovation can be given as
\begin{equation}
\mathbf{\delta z}_{RC,k}^{n}=-\begin{pmatrix}
0 & 1 & 0\\ 
0 & 0 & 1
\end{pmatrix} (\mathbf{C}_{n}^{b} \mathbf{v}_{eb}^{n} -\mathbf{\omega}_{ib}^{b} \times \mathbf{L}_{rb}^{b})_{k}
\end{equation}
where $\mathbf{C}_{b}^{n}$ is the coordinate transformation matrix from the body frame to the locally level frame, $\mathbf{L}_{rb}^{b}$ is rear wheel to body lever arm, and $\mathbf{\omega}_{ib}^{b}$ is angular rate measurement. 

\subsubsection{Wheel Odometry}
In the CoreNav method, the \ac{WO} measurements are used as an aiding state correction in the error-state \ac{EKF}. Following the same procedure as described in \cite{kilic2019improved}, the measurement innovation for the wheel odometry estimation can be given as  
\begin{equation}
\delta \mathbf{z}_{O}=\begin{pmatrix}
{\tilde{v}}_{lon,O}- {\tilde{v}}_{lon,i}\\ -{\tilde{v}}_{lat,i}\\-{\tilde{v}}_{ver,i}\\ 
\tilde{\dot{{\psi}}}_{nb,o}-\tilde{\dot{{\psi}}}_{nb,i}\overline{cos{\hat{\theta}_{nb}}}
\end{pmatrix}
\end{equation}
\begin{equation}
\begin{aligned}
\begin{bmatrix}
{\tilde{v}}_{lat,i}\\ {\tilde{v}}_{lon,i}\\ {\tilde{v}}_{ver,i}
\end{bmatrix}=&\frac{1}{\tau_0}\int_{t-\tau_{0}}^{t} \mathbf{I}_{3}\big(\mathbf{C}_{n}^{b}(\tau) \mathbf{v}_{eb}^{n}(\tau) +\mathbf{w}_{eb}^{b}(\tau)  \times \mathbf{L}_{br}^{b}\big)d\tau
\end{aligned}
\end{equation}
where $\tilde{v}_{lon}$, $\tilde{v}_{lat}$, and $\tilde{v}_{ver}$ are estimated longitudinal, lateral and vertical wheel speed, respectively. The subscript $i$ denotes the \ac{INS} estimation, and $O$ denotes the \ac{WO} measurements.

\subsection{GNSS/WO/INS Integration with ZUPT}
\label{fusion}

This section presents two factor graph strategies: 1) utilizing only the zero velocity information in a factor graph method, 2) leveraging the CoreNav position estimates (which are improved with ZUPT and non-holonomic constraints) in a factor graph method. We used the GTSAM library for graph optimization and the standard squared loss function~(L2).

\begin{figure}[h]
\centering
\includegraphics[width=\linewidth]{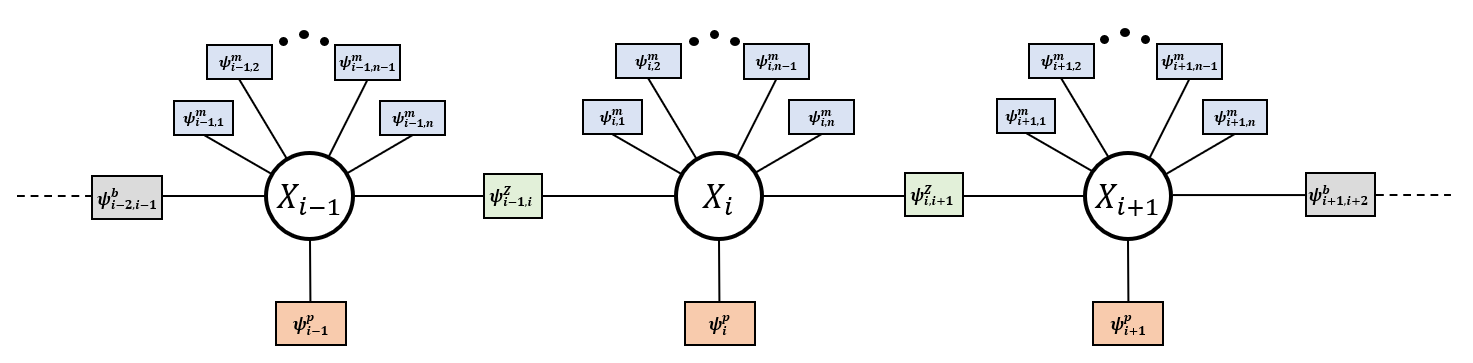}
\caption{GNSS factor graph with ZUPT factors, L2-ZUPT }
\label{fig:zuptfactorgraph}
\end{figure}

The first method does not use \ac{INS} state estimates in the factor graph directly. It only uses zero velocity signals when \ac{INS} detects the rover has stopped. To do this, a high certainty zero displacement between-factor $\psi_{b,j}$, referred to in the Fig.~\ref{fig:zuptfactorgraph} as $\psi^z$, is added between the two adjacent state vertices, which are known to be stationary. A low certainty zero-displacement between-factor is also added between non-\ac{ZUPT} vertices, where the rover is moving, to represent process noise and let the \ac{ZUPT} information propagate throughout the graph and improve the overall solution. The $\psi^z$ factors are also referred to as the ZUPT factors in the results section. An illustration of this first method is given in Fig.~\ref{fig:zuptfactorgraph} and called L2-ZUPT.

\begin{figure}[h]
\centering
\includegraphics[width=\linewidth]{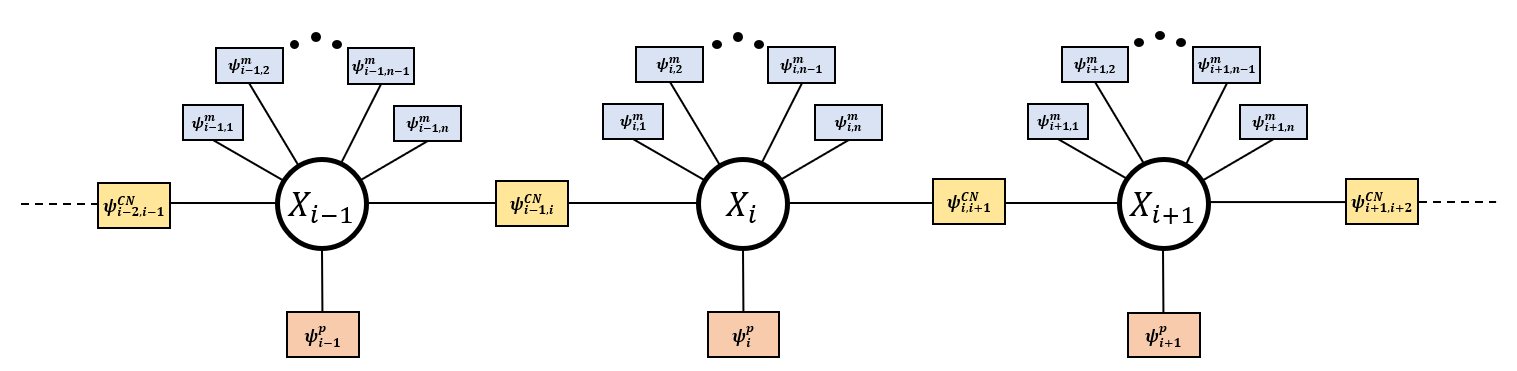}
\caption{GNSS factor graph with CoreNav factors, L2-CN. }
\label{fig:cnfactorgraph}
\end{figure}

The second method has a more direct coupling between the \ac{GNSS} and the \ac{INS}/\ac{WO} part. Here instead of utilizing the zero velocity information directly in the factor graph, the positioning solution from the CoreNav error state \ac{EKF} sensor fusion method following \cite{kilic2019improved}(see Fig.~\ref{fig:corenav}) is used. Note that, in the CoreNav method, the zero velocity information is used as a \ac{ZUPT} and also includes the non-holonomic constraints to improve the localization solution further. To couple the EKF estimates with the GNSS factor graph, the obtained positioning estimates from the CoreNav method are added as between-factors $\psi^{CN}$ among all state vertices. The $\psi^{CN}$ factors are referred to as the CoreNav factors. A depiction of the second method is given in Fig.~\ref{fig:cnfactorgraph} and called L2-CN.

\section{Experimental Results}
\label{experimental_results}

\subsection{Robot System Design}
\label{system_design}

\begin{figure}[htb!]
\centering
\includegraphics[width=0.80\linewidth]{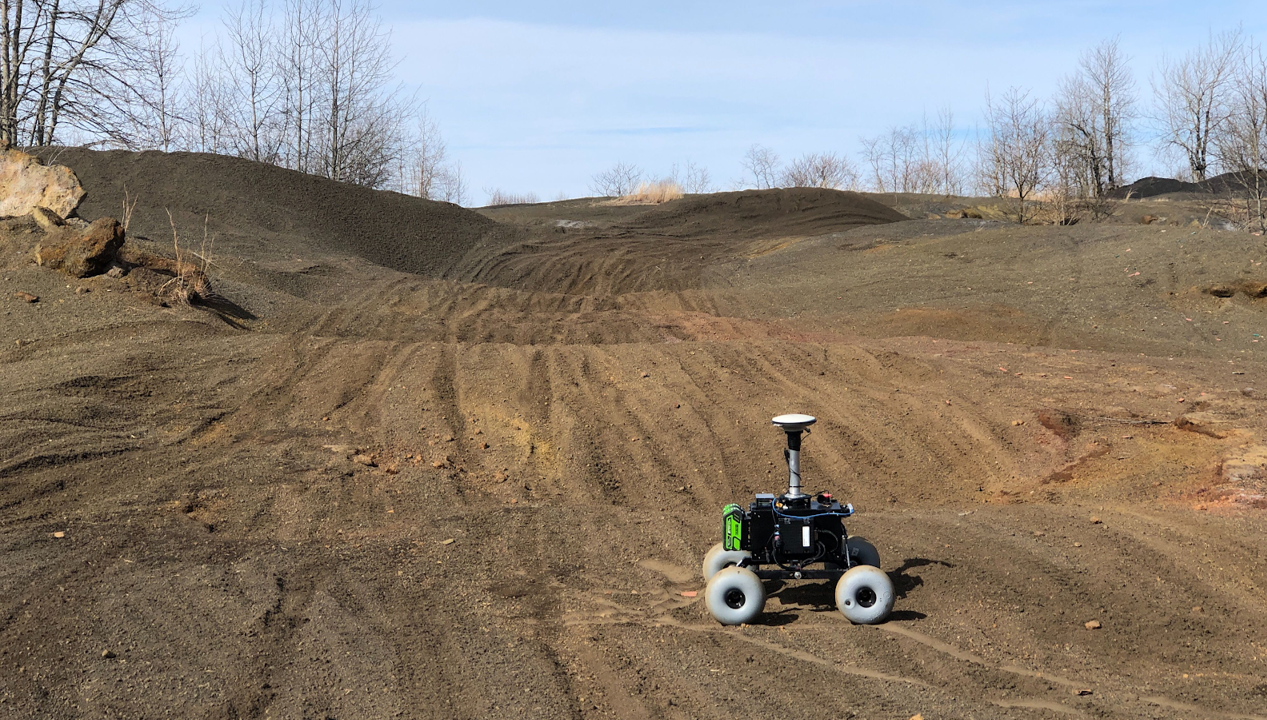}

\caption{A small wheeled rover named Pathfinder, in an off-road environment located at Point Marion, PA, where the data were obtained. }
\label{fig:pf1}
\end{figure}

In order to collect the data for validating the method, the Pathfinder testbed rover~\cite{kilic2019improved} is used. The robot is a skid-steered four-wheeled platform that can be utilized as a testing platform. A depiction of the rover is shown in Fig.~\ref{fig:pf1}. The localization sensor suite setup includes a Novatel pinwheel L1/L2 GNSS antenna~\cite{novatel2}, and receiver\cite{novatel1}, an Analog Devices ADIS-16495 \ac{IMU}~\cite{adis}, and quadrature wheel encoders. The robot is also equipped with an Intel RealSense T265 camera~\cite{intelt265}; however, this sensor is not used in the localization solution. The computer is an Intel NUC Board NUC7i7DN~\cite{intelNUC} which hosts an i7-8650U processor. The software runs on the robot under Robot Operating System (ROS)~\cite{quigley2009ros}.  

\subsection{Field Test Datasets}
\begin{figure}[h]
\centering
\includegraphics[width=0.68\linewidth]{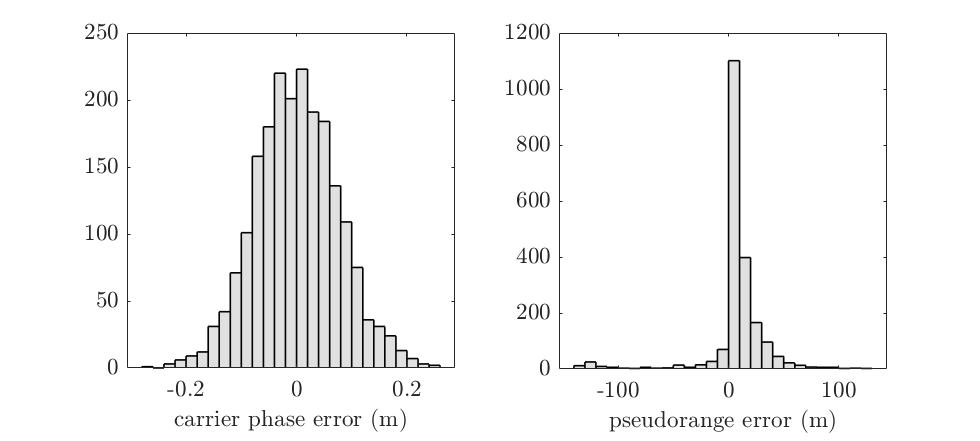}
\caption{Histogram of simulated multipath phase and range errors\cite{liu2009tracking} }
\label{fig:multipath}
\end{figure}

To evaluate the methods, we use three datasets from \cite{vz7z-jc84-20} referred in the paper as Test~1 (\texttt{ashpile\_mars\_analog1.zip}), Test~2 (\texttt{ashpile\_mars\_analog2.zip}) and Test~3 (\texttt{ashpile\_mars\_analog3.zip}). We also generate a noisy version of these datasets by adding simulated multipath noise based on the elevation of the satellites following the early-minus-late discriminator formulation in~\cite{liu2009tracking}. The noise is randomly added to 2~\% of the data in each dataset to create the noisy versions. Histogram plots of the simulated range and phase noises are shown in Fig.~\ref{fig:multipath}. The rover stops nine times for Test 1, 19 times for Test 2, and 20 times for Test 3 to obtain the zero velocity information. The distances covered in Test 1, 2, and 3 are 671m, 652m, and 663m, respectively. The reference position solutions are obtained by integer-ambiguity-fixed carrier-phase differential GPS (DGPS) processed with RTKLIB~\cite{rtklib}.

\subsection{Evaluation and Discussion}
\begin{figure}[htb!]
\centering
\includegraphics[width=0.96\linewidth]{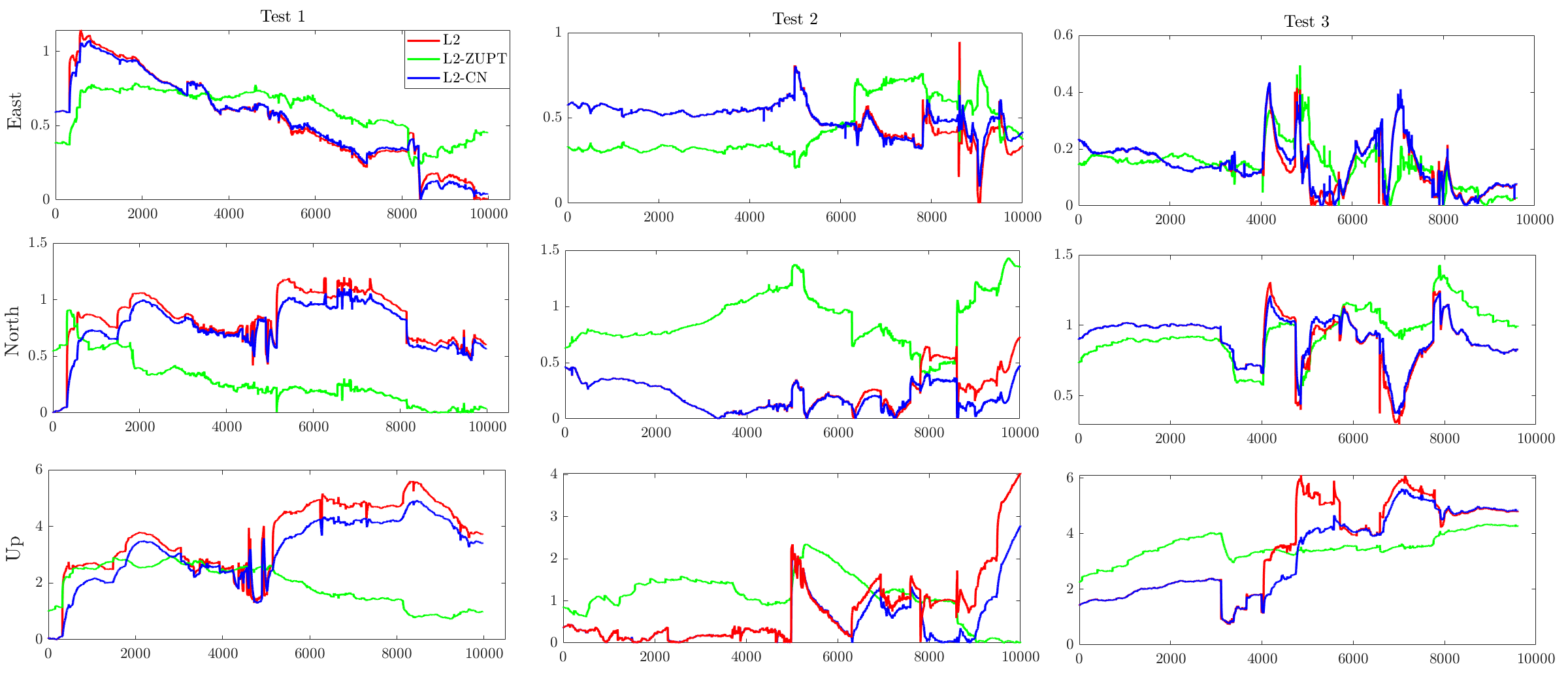}

\caption{Time variation of the errors (m) in the East-North-Up frame for clean datasets. }
\label{fig:resultENUclean}
\end{figure}

\begin{figure}[htb!]
\centering
\includegraphics[width=0.98\linewidth]{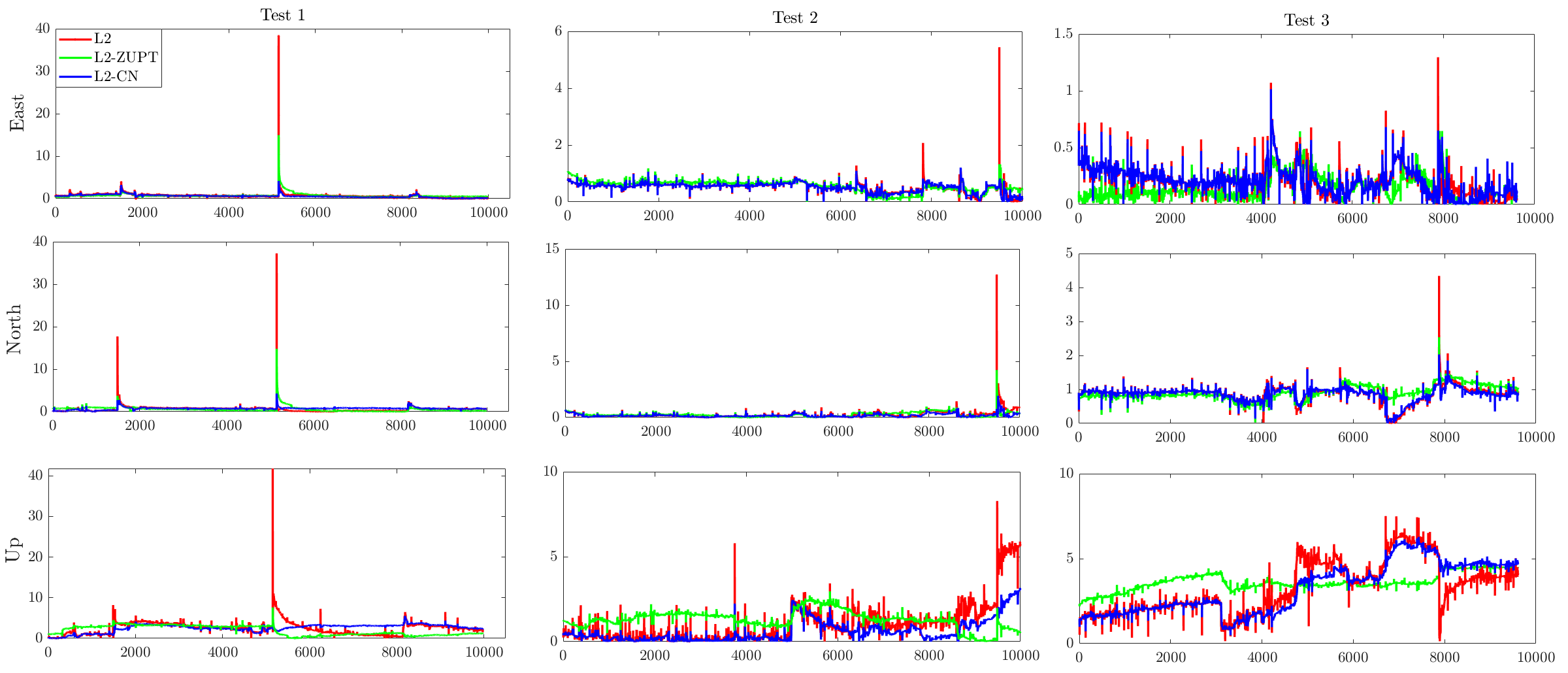}

\caption{Time variation of the errors (m) in the East-North-Up frame for noisy datasets. }
\label{fig:resultENU}
\end{figure}

The three methods compared here are the standard \ac{GNSS} factor graph (Fig.~\ref{fig:factorgraph}), the GNSS factor graph with ZUPT factors (Fig.~\ref{fig:zuptfactorgraph}) and the GNSS factor graph with CoreNav factors (Fig.~\ref{fig:cnfactorgraph}). These are referred to in the figures and tables as L2, L2-ZUPT, and L2-CN, respectively. A comparison of the methods for the clean version of the datasets is given in Table~\ref{tab:results1}. The comparison for the noisy datasets is shown in Table~\ref{tab:results2}. Figures~\ref{fig:resultENUclean} and~\ref{fig:resultENU} show the time variation of the errors in the East-North-Up frame. The larger peaks in Fig.~\ref{fig:resultENU} in the standard factor graph results are due to the large simulated multipath noise (see Fig.~\ref{fig:multipath}.)

\vspace{10 pt}
\begin{table} [htb]
\centering
\footnotesize
\begin{threeparttable}
\caption{Comparison of the methods for the clean datasets.}
\label{tab:results1}
\centering
\begin{tabular}{@{}llccccc@{}}
\hline
\multicolumn{2}{c}{Clean Dataset} & \multicolumn{4}{c}{RMSE (m)} & \multicolumn{1}{c}{Max Norm Error (m)}   \\
 & & \scriptsize{E}& \scriptsize{N}& \scriptsize{U}& \scriptsize{3D}& \scriptsize{3D}\\ 
\hline\hline
       & L2	        &0.62	&0.87	&3.82	&3.97	&5.65\\
Test 1 & L2-ZUPT	&0.62	&\colorbox{green}{0.34}	&\colorbox{green}{2.09}	&\colorbox{green}{2.20}	&\colorbox{green}{3.06}	\\
& L2-CN	            &0.62	&0.78	&3.34	&3.49	&4.97	\\
\hline
& L2	            &0.49	&0.31	&1.30	&1.43	&4.31	\\
Test 2 & L2-ZUPT	&\colorbox{green}{0.46}	&0.93	&1.24	&1.62	&\colorbox{green}{2.62}	\\
& L2-CN	            &0.51	&\colorbox{green}{0.24}	&\colorbox{green}{0.90}	&\colorbox{green}{1.06}	&3.08	\\
\hline
& L2	            &0.16	&0.92	&3.86	&3.97	&6.14	\\
Test 3 & L2-ZUPT	&0.16	&0.96	&\colorbox{green}{3.55}	&\colorbox{green}{3.69}	&\colorbox{green}{4.47}	\\
& L2-CN	            &0.16	&0.92	&3.58	&3.70	&5.65	\\
\hline\hline
% & STD	(m)        &64.14	&4.76	&3.42	&66.83	&4.97	\\
%  & Average (m)	        &85.50	&6.74	&\textbf{3.38}	&137.57	&4.70\\
% & Median	(m)        &79.39	&4.57	&\textbf{2.00}	&152.46	&3.11\\

% \hline
\end{tabular}
\begin{tablenotes}
\item[*] The best results marked up with green boxes 
\end{tablenotes}
\end{threeparttable}
\end{table}

\begin{table} [htb]
\centering
\footnotesize
\begin{threeparttable}
\caption{Comparison of the methods for noisy datasets.}
\label{tab:results2}
\centering
\begin{tabular}{@{}llccccc@{}}
\hline
\multicolumn{2}{c}{Noisy Dataset} & \multicolumn{4}{c}{RMSE (m)} & \multicolumn{1}{c}{Max Norm Error (m)}   \\
 & & \scriptsize{E}& \scriptsize{N}& \scriptsize{U}& \scriptsize{3D}& \scriptsize{3D}\\ 
\hline\hline
 & L2	            &0.88	&0.87	&2.85	&3.10	&67.81  \\
Test 1 & L2-ZUPT	&0.88	&0.78	&\colorbox{green}{2.27}	&\colorbox{green}{2.55}	&22.39	\\
& L2-CN	            &\colorbox{green}{0.67}	&\colorbox{green}{0.77}	&2.72	&2.90	&\colorbox{green}{7.29}	\\
\hline
& L2	            &0.53	&0.40	&1.77	&1.90	&16.12	\\
Test 2 & L2-ZUPT	&0.59	&0.33	&1.36	&1.52	&4.85	\\
& L2-CN	            &0.53	&\colorbox{green}{0.24}	&\colorbox{green}{1.01}	&\colorbox{green}{1.16}	&\colorbox{green}{3.41}	\\
\hline
& L2	            &0.23	&0.90	&3.59	&3.71	&7.52	\\
Test 3 & L2-ZUPT	&\colorbox{green}{0.17}	&0.93	&3.65	&3.77	&\colorbox{green}{5.08}	\\
& L2-CN	            &0.23	&\colorbox{green}{0.88}	&\colorbox{green}{3.55}	&\colorbox{green}{3.66}	&6.33	\\
\hline\hline
% & STD	(m)        &64.14	&4.76	&3.42	&66.83	&4.97	\\
%  & Average (m)	        &85.50	&6.74	&\textbf{3.38}	&137.57	&4.70\\
% & Median	(m)        &79.39	&4.57	&\textbf{2.00}	&152.46	&3.11\\

% \hline
\end{tabular}
% \begin{tablenotes}
% \item[*] WIO: Wheel Odometry + IMU Heading, VO: Visual Odometry, \\VIWO: VO Velocity + IMU + WO Velocity.
% \end{tablenotes}
\end{threeparttable}
\end{table}

\newpage

L2-ZUPT and L2-CN have better performance than the GNSS-only factor graph, L2, for clean and noisy datasets 1 and 2. We see comparable performances from all three methods for dataset 3. L2-ZUPT is found to perform best for clean data, and L2-CN performs best for noisy data.  The effect of leveraging zero velocity information can be seen in the noisy data results, where the larger errors in the standard factor graph are dampened by constraining the states during the ZUPTs to be the same since the rover is stationary at that time. The better performance of L2-CN for noisy data can be explained by the fact that the factor graph utilizes more information since it uses the IMU-WO solution from CoreNav. CoreNav also uses additional non-holonomic constraints which are not used in the L2 and L2-ZUPT methods. The performance gap between L2-ZUPT and L2-CN is affected by the number of stops in the datasets. For example, we see a more significant performance gap between L2-ZUPT and L2-CN in Test 1, which has fewer stops. This performance gap is smaller in Test 2, and 3 where the numbers of stops are more than in Test 1, which indicates that using only ZUPT factors in the GNSS factor graph can provide similar localization performance as using a GNSS/WO/INS coupled solution.

\section{Conclusions and Future Work}
\label{conclusion}
This work presents GNSS/WO/INS fusion strategies with a factor graph by exploiting the zero velocity information to improve the positioning solution for wheeled robots. The presented methods are compared with a standard GNSS factor graph. This comparison shows that the zero velocity information is a valuable constraint to further improving the positioning solution in a GNSS-degraded environment. Moreover, we observed that using only zero velocity information in a GNSS factor graph can provide comparable positioning performance as using a GNSS/INS/WO coupled position solution. To evaluate our \ac{ZUPT} aided factor graph methods, we used open access datasets~\cite{vz7z-jc84-20}. Also, we made our software implementation publicly available. 

Future works will involve testing with more extended datasets with real multipath noise and investigating the connection with robust filtering algorithms. We also experimented with the incremental covariance estimation method discussed in \cite{watson2020robust}. The ZUPT factors are expected to help learn the measurement noise covariance model better. However, the parameter tuning of incremental covariance estimation and ZUPT became a challenge, and due to this, we could not find consistent results with these datasets, which could be attributed to the shorter length of the trajectories.

\section*{Acknowledgements}
This work was supported in part by NASA EPSCoR Research Cooperative Agreement WV-80NSSC17M0053, and in part by the Benjamin M. Statler Fellowship. 

\bibliographystyle{IEEEtran}
\bibliography{references}

\end{document}